\documentclass{article}

\usepackage[preprint]{neurips_2024}
\usepackage[utf8]{inputenc} 
\usepackage[T1]{fontenc}    
\usepackage{hyperref}       
\usepackage{url}            
\usepackage{booktabs}       
\usepackage{amsfonts}       
\usepackage{nicefrac}       
\usepackage{microtype}      
\usepackage{natbib}
\usepackage{amsmath}
\usepackage{amsfonts}
\usepackage{amsthm}
\usepackage{booktabs}
\usepackage{algorithm}
\usepackage{algorithmic}
\usepackage{amssymb}
\usepackage{graphicx,wrapfig}
\usepackage{booktabs}
\usepackage{multirow}
\usepackage{bbm}
\usepackage{makecell}
\usepackage[table,xcdraw,dvipsnames]{xcolor}
\usepackage{marvosym}
\usepackage{soul}
\usepackage{ulem}
\usepackage{comment}
\usepackage{listings}  
\usepackage{mylstlinebgrd}
\setcitestyle{numbers,square}
\usepackage[table]{xcolor}

\definecolor{eclipseRed}{RGB}{255,0.0,0}
\definecolor{eclipseBlue}{RGB}{42,0.0,255}
\definecolor{eclipseGreen}{RGB}{63,180,95}
\definecolor{eclipsePurple}{RGB}{175,0,25}
\definecolor{codewhite}{rgb}{0.70,0.70,0.70}
\definecolor{codegray}{rgb}{0.5,0.5,0.5}
\definecolor{codepurple}{rgb}{0.58,0,0.82}
\definecolor{backcolour}{rgb}{0.95,0.95,0.92}

\newcommand{\ziyue}[1]{\textcolor{blue}{\textit{[\textbf{Ziyue}: #1]}}}
\newcommand{\re}[1]{{\color{black}#1}}

\definecolor{instructioncolor}{rgb}{0.6275, 0.7686, 0.6157}
\definecolor{taskcolor}{rgb}{0.8824, 0.9255, 0.7843}
\definecolor{prefixcolor}{rgb}{0.9922, 1, 0.6824}
\lstdefinelanguage{Prompt}{
	backgroundcolor=\color{backcolour},   
	keywordstyle=\underbar,
	numberstyle=\tiny\color{codegray},
	basicstyle=\ttfamily\footnotesize,
	breakatwhitespace=false,         
	breaklines=true,   
    breakindent=-5pt,
    moredelim=[is][\sout]{|}{|},
	captionpos=b,                    
	keepspaces=true,                 
	numbers=left,                    
	numbersep=5pt,                  
	showspaces=false,                
	showstringspaces=false,
	showtabs=false,                  
	tabsize=4,
	escapeinside={`}{`},
        morecomment = [s][\color{eclipseRed}\bfseries]{\$}{\}},
	morecomment = [s][\color{eclipseGreen}\bfseries]{How}{?},
        morecomment = [l][\color{eclipseBlue}\bfseries]{SELECT},
        morecomment = [l][\color{eclipsePurple}\bfseries]{\$\{DATABASE_SCHEMA\}},
        morecomment = [s][\color{eclipsePurple}\bfseries]{CREATE}{;},
        morecomment = [l][\color{eclipsePurple}\bfseries]{Table},
        morecomment = [l][\color{eclipsePurple}\bfseries]{stadium},
        morecomment = [l][\color{eclipsePurple}\bfseries]{singer},
        morecomment = [l][\color{eclipsePurple}\bfseries]{concert},
        morecomment = [l][\color{eclipsePurple}\bfseries]{singer_in_concert},
        morecomment = [l][\color{codewhite}\bfseries]{\$\{TARGET_QUESTION\}},
    postbreak={
       \mbox{
           \lst@linebreakbgrd
           \rotatebox[y=0.7ex]{180}{\color{black}$\Lsh\,$}
       }
    },
}

\lstdefinelanguage{Special_Prompt}{
	backgroundcolor=\color{backcolour},   
	keywordstyle=\color{magenta},
	numberstyle=\tiny\color{codegray},
	basicstyle=\ttfamily\footnotesize,
	breakatwhitespace=false,         
	breaklines=true,   
    breakindent=-5pt,
	captionpos=b,                    
	keepspaces=true,                 
	numbers=left,                    
	numbersep=5pt,                  
	showspaces=false,                
	showstringspaces=false,
	showtabs=false,                  
	tabsize=4,
	escapeinside={`}{`},
        moredelim=**[is][\color{eclipsePurple}\bfseries]{\&\&}{\&\&},
        moredelim=**[is][\color{eclipseGreen}\bfseries]{\^\^}{\^\^},
        moredelim=**[is][\color{eclipseBlue}\bfseries]{@@}{@@}, 
    postbreak={
       \mbox{
           \lst@linebreakbgrd
           \rotatebox[y=0.7ex]{180}{\color{black}$\Lsh\,$}
       }
    },
}

\lstdefinelanguage{Question}{
	backgroundcolor=\color{backcolour},   
	sensitive = true,
	morecomment = [s]{Gold}{SQL},
	commentstyle ={\color{red}\bfseries\underbar},
	morestring = [b]",
	morestring = [b]',
	stringstyle = \color{eclipseGreen},
	basicstyle=\ttfamily\footnotesize,
	breaklines=true,
	alsoletter=!?-,
	emph={StableVicuna-13B, Vicuna-13B, Gold-SQL, Write a sql to answer},
	emphstyle={\color{red}\bfseries\underbar},
	captionpos=b,
	escapeinside={[}{]},
	keepspaces=true,              
	showspaces=false,                
	showstringspaces=false,
	showtabs=false,                  
	tabsize=4,
	columns=flexible
}

\title{\texttt{PET-SQL}: A Prompt-Enhanced Two-Round Refinement of Text-to-SQL with Cross-consistency
}
\author{%
Zhishuai Li$^{1,*}$, Xiang Wang$^{1,*}$, Jingjing Zhao$^{1,*}$, Sun Yang$^{2,*}$, Guoqing Du$^{1,*}$,\\ \textbf{Xiaoru Hu$^{1,*}$, Bin Zhang$^{1,3,4,*}$, Yuxiao Ye$^{1,5,}$\thanks{Equal contribution.},\quad
  Ziyue Li$^{1,6,}$\textsuperscript{\Letter}, 
  Rui Zhao$^{1,}$\textsuperscript{\Letter}, 
  Hangyu Mao$^{1}$}
  \\
  $^1$SenseTime Research\\
    $^2$Peking University\\
  $^3$Institute of Automation, Chinese Academy of Sciences\\
  $^4$School of Artificial Intelligence, University of Chinese Academy of Sciences\\
  $^5$School of Computer Science and Technology, Beijing Institute of Technology\\
  $^6$University of Cologne, Germany
}

\begin{document}
\maketitle
\def\thefootnote{$^\textrm{\Letter}$}\footnotetext{Corresponding author: \{{liziyue,zhaorui}@sensetime.com\}}\def\thefootnote{\arabic{footnote}}

\begin{abstract}
Recent advancements in Text-to-SQL (Text2SQL) emphasize stimulating the large language models (LLM) on in-context learning, achieving significant results. Nevertheless, they face challenges when dealing with verbose database information and complex user intentions. This paper presents a two-round framework to enhance the performance of current LLM-based natural language to SQL systems. We first introduce a novel prompt representation, called reference-enhanced representation, which includes schema information and randomly sampled cell values from tables to instruct LLMs in generating SQL queries. Then, in the first round, question-SQL pairs are retrieved as few-shot demonstrations, prompting the LLM to generate a preliminary SQL (\texttt{PreSQL}). After that, the mentioned entities in \texttt{PreSQL} are parsed to conduct schema linking, which can significantly compact the useful information. In the second round, with the linked schema, we simplify the prompt's schema information and instruct the LLM to produce the final SQL (\texttt{FinSQL}). Finally, as the post-refinement module, we propose using cross-consistency across different LLMs rather than self-consistency within a particular LLM. Our methods achieve new SOTA results on the Spider benchmark, with an execution accuracy of 87.6\%. The codes are released on \url{https://github.com/zhshLii/PETSQL}
\end{abstract}


\section{Introduction}
Text-to-SQL (Text2SQL) assists individuals in converting natural language questions (i.e., text) into structure query language (SQL) queries. 
For example, conditioned on the contents of the user's question ``\texttt{How many singers do we have?}'' and the corresponding description for the database schema (e.g., table/column names), the ideal Text2SQL agent can generate SQL statements ``\texttt{SELECT count(*) FROM singer}'', which can then be executed on the SQL database engine to retrieve the desired response.
The task is helpful for the database question-answering and information-retrieval applications, such as finance, traffic, and business intelligence \cite{sui2023reboost,zhang2024benchmarking}, as it lowers the requirements for expertise and allows non-professional participants to interact with databases in natural language.

\begin{figure}
    \centering
    \includegraphics[width=0.9\textwidth]{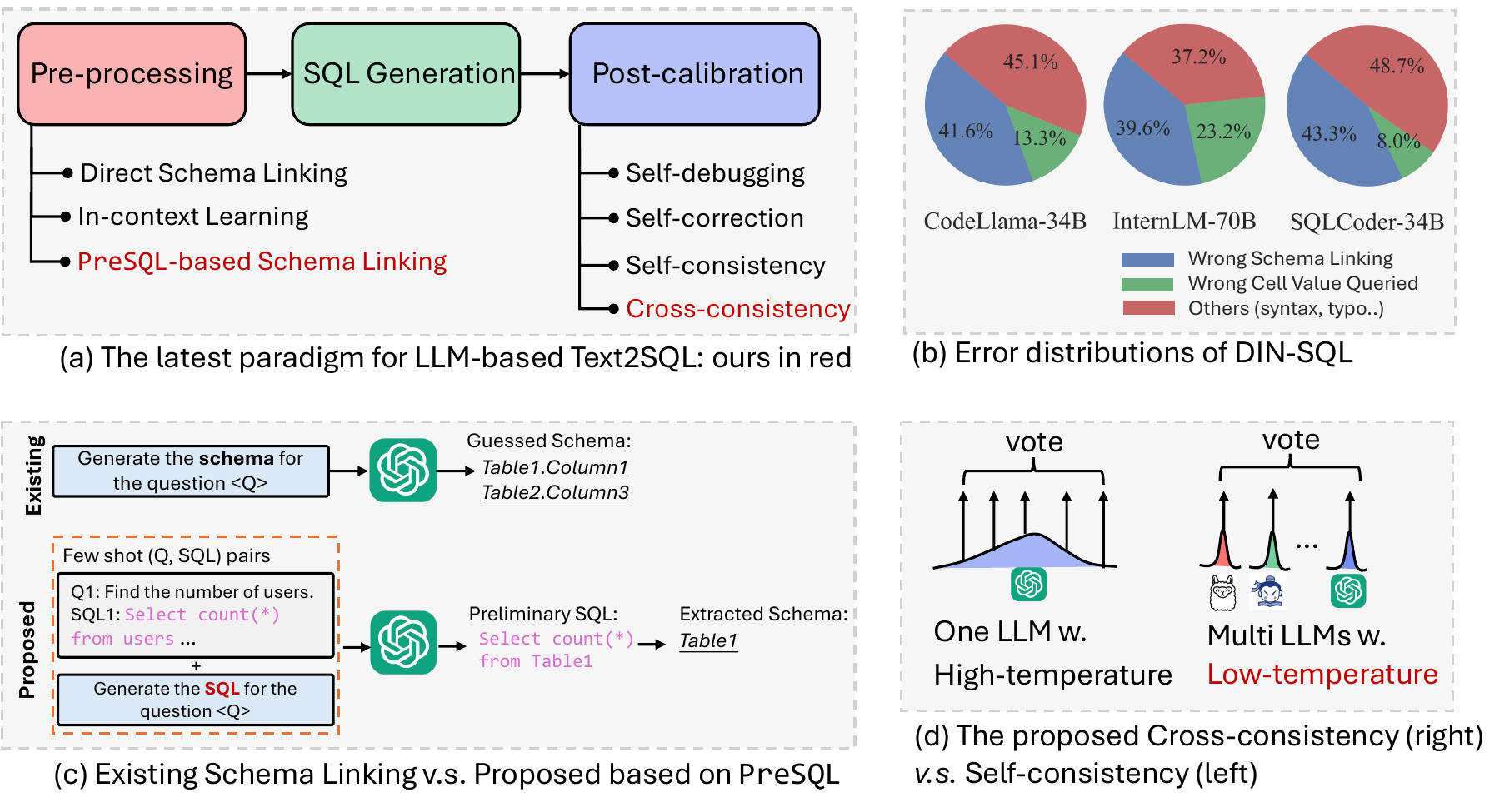}
    \vspace{-10pt}
    \caption{(a) A three-stage paradigm of SOTA Text2SQL. (b) We summarize the distribution of error types in DIN-SQL \cite{din-sql}. We elaborate on the methods for each type: \texttt{PreSQL}-based schema linking for errors in schema linking, cell values references-enhanced prompt for incorrect cell values, and cross-consistency for other errors caused by typos or syntax mistakes. (c) \re{Existing schema linking directly lets LLM guess the schema, we let LLM first generate a preliminary SQL and parse table/column entities from \texttt{PreSQL}, increasing accuracy by \textbf{+4\%}}. (d) Our proposed cross-consistency enables the LLMs to operate at low temperatures, thereby reducing hallucination.
    }
    \label{fig:intro}
    \vspace{-10pt}
\end{figure}

\re{The development trajectory of Text2SQL has gone from rule-based, domain-specific solutions to deep learning-based models, specifically sequence-to-sequence models. 
Recent large language models (LLM)-based solutions have further pushed the performance to a new high. The recent state-of-the-art high-performers \cite{din-sql,dail-sql} have decomposed the Text2SQL into a series of subtasks. Such a task decomposition has become a new paradigm, and we generalize it as three stages, as shown in Fig. \ref{fig:intro}(a): (1) \textbf{pre-processing}, including prompt engineering such as zero-shot \cite{c3} or few-shot prompting \cite{din-sql}, schema linking \cite{din-sql,yang2024sqltoschema}, in-context-learning (ICL) \cite{din-sql}, (2) \textbf{SQL generation}, which is based on LLMs such as CodeLlaMa, ChatGPT, GPT-4, (3) \textbf{post-calibration}, including self-consistency \cite{dail-sql}, self-debugging \cite{sui2023reboost,wang2023mac}, self-correction \cite{din-sql}. Most of the improvements and designs are happening in the preprocessing and post-calibration, particularly in schema linking (+4\% in DIN-SQL with CodeX Davinci) and self-consistency (+0.4\% in DAIL-SQL with GPT-4). 


However, when handling extensive databases across various domains with different formats and intricate user intentions, there is still large room for improvement in preprocessing and post-calibration. 

According to the statistics on the Spider \cite{spider}, an opensource benchmark database for Text2SQL with 10,181 questions and 
across 200 complex databases in 138 domains with multiple tables, covering almost all important SQL components including GROUP BY, ORDER BY, HAVING and nested queries,
around 60.7\% questions are about ``\texttt{WHERE =}'' or  ``\texttt{HAVING =}''. However, due to various cell values being formatted, an SQL can be semantically correct but executively wrong. Take a SQL question of querying gender as an example, male can be formatted variously as `\textit{sex=`Male'}', `\textit{sex=`male'}', or `\textit{sex=`M'}', and there are more examples from our experiences, such as 1000 also being `1k' or `1000.0': Such a formatting variance is even more commonplace and critical in real industrial databases, and misaligned formatting in SQL will produce different results or even fail. 
Thus, in the prompting, we proposed a \textbf{cell value reference}, which provides standardized references in filling out different tables and helps LLM understand the formats and specifications of the database, and in our experiment, it can improve the execution accuracy by maximum \textbf{+4\%}.

Furthermore, schema-linking, indicating and narrowing down to which tables and which columns the LLMs should search for the answer, is a critical sub-task and performance-booster for Text2SQL. Although works like DIN-SQL \cite{din-sql} proved that directly using LLMs to infer the tables and columns of interest can work better than an end-to-end trained solution, such as BERT \cite{devlin2018bert}, we claim that LLMs even GPT-4, have \textbf{NOT} been trained or supervised fine-tuned with such a domain-specific schema linking task (i.e., input a specific SQL question and output the relevant table names and column names), as it falls outside the mainstream scope of LLMs.
As a result, the schema linking still accounts for about 22\% of the mistakes in GPT-4, while the conditions in other LLMs are more serious (see Fig. \ref{fig:intro}(b)). We innovatively proposed a preliminary-SQL (\texttt{PreSQL})-based schema linking, in which we first generate a rough SQL statement by the LLMs and then parse the mentioned table/column entities as the linking results (see Fig. \ref{fig:intro}(c)). 
The rationale behind this is that code (particularly SQL) generation, is a fundamental task across nearly all LLMs, and they may be adequately pre-trained in relevant corpus. Therefore, for LLMs, generating \texttt{PreSQL} is easier and more natural than directly performing schema linking as instructed.
Consequently, our \texttt{PreSQL}-based schema linking achieves maximum \textbf{+4\%} higher execution accuracy.

Last but not least, two major designs in post-calibration, self-debugging and self-consistency, either have very trivial effects or create more instability in the results, according to our findings. According to our experiences with multiple LLMs, we found that self-debugging is useful for fixing syntax errors, but it tends to be powerless in front of the SQLs generated by strong LLMs, whose problem is semantic ambiguity rather than syntax errors. 
In terms of self-consistency, the technique that is proposed by the current top-1 opensource solution, is theoretically yielding suboptimal SQL due to its nature: it achieves self-consistency by calling the same LLM several times at a \textbf{higher temperature} and choosing the majority as the final output, which firstly does not fundamentally diversify the SQL outputs, and moreover, the higher temperature generates more hallucinations, potentially yielding a stable but bad output. We instead innovatively propose a ``cross-consistency'' (in Fig. \ref{fig:intro}(d)), which calls multiple different LLMs at a low temperature, guaranteeing diversity (due to different LLMs) and further converging to a high-quality result (due to low temperature). We also further proposed two voting schemes, which are (1) native majority voting of the executed results and (2) difficulty-aware voting with distinct candidate LLMs according to the complexity grades of \texttt{PreSQL}.
Overall, the method achieves state-of-the-art in the Spider  \cite{spider} leaderboard with 87.6\% execution accuracy, being ranked as the top-1 opensource solution. 
}
The main contributions are threefold:
\begin{itemize}
    \item An elaborated prompt is designed and its effectiveness is systemically evaluated under various LLMs. Besides the schema information, the cell values in tables and magic instructions are attended to as the prior knowledge in the prompt.
    \item We explore the \texttt{PreSQL}-based schema linking instead of prompting the LLM to directly generate table/column names. We believe it can yield more concise results and is suitable for coding LLM. Coupled with linked schema, we present to organize the simplified prompt and then feed it into the LLM again. 
    \item The cross-consistency is proposed to utilize the diversity across different LLMs. The \texttt{PreSQL} is also re-used to vote in this module. Overall, by voting across various LLMs, the final predicted SQL achieves 87.6\% execution accuracy in the Spider benchmark.
\end{itemize}
\section{Related Work}
\textbf{Learning-based agents.}
With the advancement of language models in previous years, multiple methods have been proposed for the Text2SQL task. These early approaches focus on pattern matching between natural language (NL) and SQL statements~\cite{seq2sql, sqlnet, hydranet}. Some studies employ relation-aware self-attention mechanisms as encoders to learn representations of questions and schemas. These methods then utilize grammar-based decoders to generate SQL as abstract syntax trees or employ sketch-based decoders to obtain SQL by filling in slots~\cite{2017syntax-tree, 2018coarse2fine}. Subsequently, the advent of pre-trained language models and the fine-tuning paradigm significantly influences these methods~\cite{T5}.  Numerous studies utilize standard sequence-to-sequence models with transformer architectures~\cite{transformer} to translate NL questions into SQL queries in an end-to-end manner~\cite{picard, SeaD}.

\textbf{LLM-based in-context learning methods.}
The development of LLMs has catalyzed substantial transformations in this field.
These methods leverage the in-context learning, semantic understanding, and reasoning capabilities of LLMs~\cite{ruan2023tptu, kong2023tptu, zhang2023controlling}, continuously pushing the boundaries of performance on various evaluation benchmarks, such as Spider~\cite{spider} and BIRD~\cite{bird}.
For example, C3~\cite{c3} achieves a breakthrough by leveraging ChatGPT's zero-shot learning capability. 
DIN-SQL ~\cite{din-sql} decomposes the task into multiple components, including schema linking, difficulty classification, and SQL generation. 
This methodology effectively reduces the complexity of decision-making, enabling LLMs to generate more precise SQL statements.
DAIL-SQL~\cite{dail-sql} further enhances the capabilities of LLMs through in-context learning and supervised fine-tuning schemes.
In addition, several other studies incorporate advanced reasoning methods, such as chain-of-thought~\cite{cot} and self-reflexion~\cite{self-reflexion}, to enhance the capabilities of LLMs and improve their performance on Text2SQL tasks~\cite{act-sql, DIR}.

\section{Methodology}\label{sec:method}
For a natural language question $Q$ on a database $D$, the objective of LLM-based Text2SQL is to translate $Q$ into an executable SQL query ${s}$. The likelihood of an LLM $\mathcal{M}$ generating a SQL query ${s}$ is defined as a conditional probability distribution, where $\mathcal{I}(D)$ is the description of $D$, 
$\mathcal{P}$ is the prompt template, and $|s|$ is the length of $s$. $s_i$ and $s_{<i}$ are the $i$-th token and the prefix of $s_i$:
\begin{equation} \small
    \mathbb{P}_{\mathcal{M}}(s|\mathcal{P}({Q}, \mathcal{I}(D))) = \prod_{i=1}^{|s|} \mathbb{P}_{\mathcal{M}}(s_i | \mathcal{P}(Q, \mathcal{I}(D)), s_{<i}),
\end{equation}

\begin{figure}
    \centering
    \includegraphics[width=\textwidth]{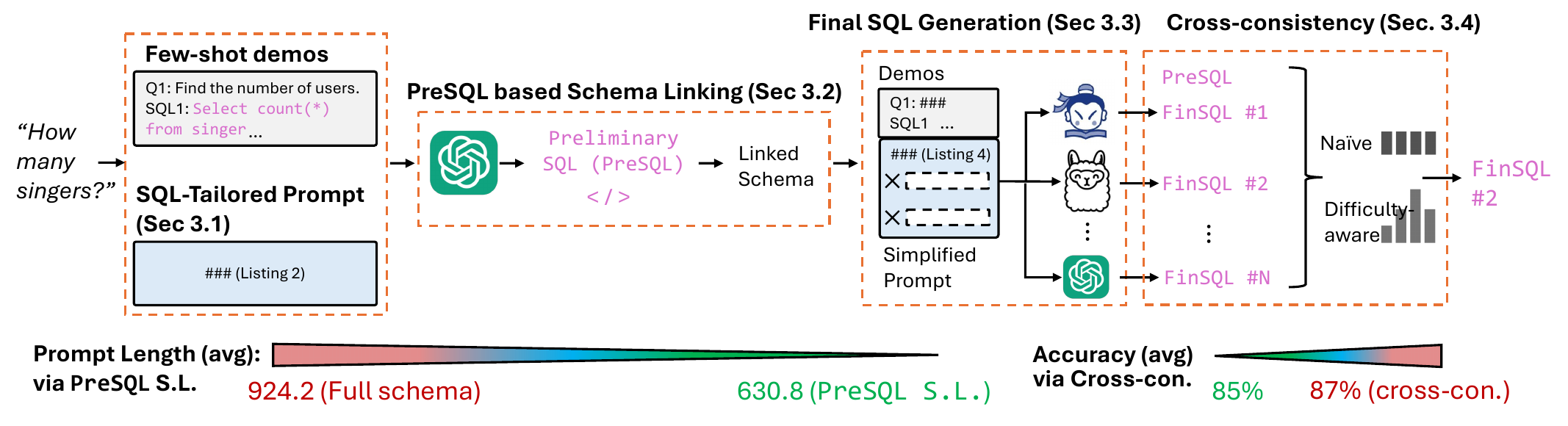}
    \vspace{-20pt}
    \caption{The overview of PET-SQL framework. It first elaborates on an effective SQL-Tailored prompt (Sec.  \ref{sec:prompt}), which can be enhanced with selected demos as few-shot in-context learning. Then, \texttt{PreSQL} is generated by LLMs for parsing linked schema (Sec.  \ref{sec:schema}), where linked entities are retained while unrelated parts are removed from the prompt, thereby simplifying it. After that, the Final SQL can be obtained from LLMs with the concise prompt (Sec.  \ref{sec:finsql}). Finally, cross-consistency is proposed to post-refine the results generated from multiple LLMs (Sec.  \ref{sec:cc}). \re{Our \texttt{PreSQL}-based schema linking shortens the prompt length by 32\%, and our cross-consistency boosts the Text2SQL accuracy by 2\%.}}
    \vspace{-10pt}
    \label{fig:arch}
\end{figure}

As shown in Fig. \ref{fig:arch}, our PET-SQL framework mainly includes: (1) an elaborated prompt that harnesses the customized instructions, basic database information, and samples in the stored tables (2) instructing the LLM to generate \texttt{PreSQL}, in which some demonstrations are selected from pools using a question similarity-based strategy and then prefixed to the prompt as the few-shot in-context (3) finding question-related tables (schema linking) based on the \texttt{PreSQL} and prompt LLM to yield FinSQL by the linked schema (4) ensuring consistency in the predicted results across multiple LLMs.

\subsection{Pre-processing: SQL-Tailored Prompting}
\label{sec:prompt}

When instructing LLMs to generate SQL queries, the usage of styles or templates for prompts significantly impacts LLMs' performance. As  \cite{dail-sql} recommended, \textbf{C}ode \textbf{R}epresentation ($CR_p$) and \textbf{O}penAI \textbf{D}emonstration ($OD_p$) forms mixed with additional information are better choices (shown in Listing 1).
To further exploit the potential of LLM, we enrich the prompt based on OpenAI Demonstration ($OD_p$), and \re{we name our proposed as \textbf{R}eference-\textbf{E}nhanced representation ($RE_p$)}.



\renewcommand{\figurename}{Listing}
\setcounter{figure}{0}
\begin{wrapfigure}{r}{0.3\textwidth}  
  \vspace{-18pt}
  \begin{center}
    \includegraphics[width=0.3\textwidth]{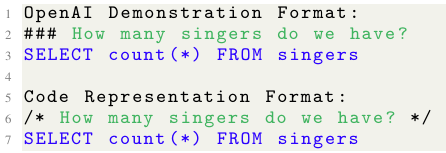} 
  \end{center}
  \vspace{-10pt}
  \caption{OpenAI Demonstration v.s. Code Representation.}
    \label{lst:infix}
    \vspace{-10pt}
\end{wrapfigure}
Building upon $OD_p$ consisting of the task instruction, database schema, and question components, \re{our proposed} $RE_p$ performs three modifications: optimization rule (OR), cell value references (CV), and foreign key declarations (FK), \re{which all turn out effective.} 

\textbf{Optimization Rule (OR)}: Concretely, in the task instruction, we raise a multi-task constraint rule for LLM, that is, highlighting the LLM to ``minimize SQL execution time while ensuring correctness'' (line 1 in Listing 2). The rationale is not only to focus on execution accuracy but also to make LLM aware of the efficiency of SQL statements. During the generation of efficient SQL statements by LLM, redundant characters and operators can be avoided, which often result in exceptions. \re{As our experiment shows, such an optimization rule can boost the performance by as high as \textbf{+2.3\%}}.
\renewcommand{\figurename}{Listing}
\setcounter{figure}{1}
\begin{figure}[h]
  \begin{center}
    \includegraphics[width=0.7\textwidth]{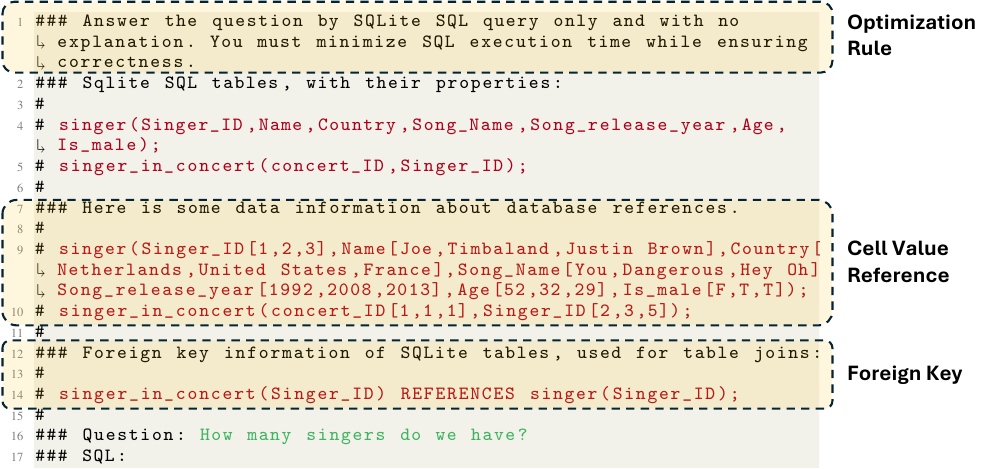} 
  \end{center}
  \caption{The proposed \re{SQL-Tailored} prompt, \re{containing OR, CV, and FK.} It is zero-shot without demos.}
    \label{list:prompt}
\end{figure}

\textbf{Cell Value References (CV)}: \re{As mentioned before, to avoid mistakes from wrong-formatted cell value,} three rows in each table are randomly sampled and inserted in the prompt (lines 7-11 in Listing 2).
The sampled cell values as references can help LLM understand the database formats and specifications. It relieves the dilemma of being unsure which element to query as the condition caused by the lack of standardization in filling out different tables. Without prior knowledge of cell values, organizing condition statements to query the number of men in a table can be confusing. \re{Such a cell value reference can boost performance by \textbf{+4.0\%}.}

\textbf{Foreign Key Declarations (FK)}: Thirdly, foreign key relations in the schema are added as suffixes in the prompt (lines 12-15 in Listing 2). It facilitates LLM in recognizing the connections between the tables involved in the database, thus improving the understanding of the user's intent and automatically selecting the appropriate connection in the queries. \re{Such foreign key can boost the performance by as high as \textbf{+7.5\%}}.

The prompt above is in a zero-shot setting and can be further enhanced by the few-shot approach. We will explain how to select demonstrations as shots, equip them with the prompt, and stimulate the LLM to yield the preliminary SQL (\texttt{PreSQL}) for a question. \re{Such a \texttt{PreSQL} is a critical design for our schema linking with higher accuracy.}

\subsection{\re{Pre-processing: To Generate A Preliminary SQL (\texttt{PreSQL}) for Schema Linking}}
\label{sec:schema}
To eschew the verbose information in the schema that may obstruct the LLMs' performance, 
we further conduct the schema linking (SL), \re{which identifies and narrows down the table and column references that are related} to the database schema and condition values related to the natural language question. 
\re{Schema link has been proven to boost the accuracy}, enhance generalizability across domains, and facilitate the synthesis of complex queries \cite{c3,din-sql}. 

\re{However, existing methods directly employed LLMs to ``guess'' the schema, and such a ``guess'' can be risky since} \textbf{LLMs, even GPT-4, have NOT been pre-trained or supervised and fine-tuned with such a domain-specific schema linking task}. 
\re{Though not being trained with schema linking, most LLMs, luckily, are trained with code generation tasks, such as SQL generation. So our intuitive yet very innovative solution is: \textbf{can we solve the schema linking task (what LLM is not good at) by letting LLM generate SQL code (what LLM is good at)?} 
}
Thus, rather than instructing the LLMs to link schema, we instruct 
LLMs to generate a preliminary SQL (\texttt{PreSQL}), and then the table and column entities mentioned in the PreSQL are parsed as the linking results. To ensure more accurate schema linking, we incorporate a few steps to generate the \texttt{PreSQL}:


\textbf{Step 1 - Question De-semanticization}\cite{guo2023case}: the domain-related tokens (i.e., table names `\texttt{singer}', column names `\texttt{gender}', and values `\texttt{male}') in questions are masked with a special token \texttt{<mask>} according to the database schema, and thus obtain the question skeletons that only exhibit the question's intentions.

\textbf{Step 2 - Demo Retrieval based on Similarity}: Then, all the question skeletons and question-SQL pairs in the training set are constructed as a demonstration pool. 
Based on the semantic embeddings of question skeletons, we retrieve the top-$K$ similar samples from the pool with the target question. Like  \cite{dail-sql}, the embedding model for question skeletons is a pre-trained sentence Transformer.

\textbf{Step 3 - Few-shot demos + SQL-Tailored Prompt}: After that, we organize the selected demonstrations with our prompt $RE_p$ following  \cite{dail-sql}, that is, prefixing question-SQL pairs as the few-shot contexts in $RE_p$ (as shown in lines 1-10 in Listing 3). Finally, the few-shot $RE_p$ is used to prompt the LLM and generate \texttt{PreSQL}.

\renewcommand{\figurename}{Listing}
\setcounter{figure}{2}
\begin{figure}[h]
  \begin{center}
    \includegraphics[width=0.7\textwidth]{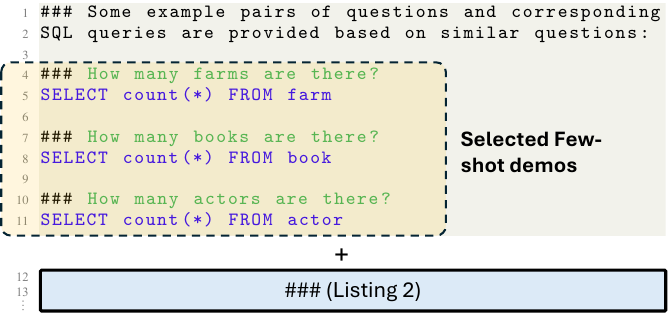} 
  \end{center}
  \caption{Three question-SQL pairs are retrieved and prefixed as a 3-shot prompt, \re{followed by Listing \ref{list:prompt}} to serve as the full prompt feeding into LLM for \texttt{PreSQL}.}
    \label{list:3shotrd}
\end{figure}






\textbf{Step 4 - Schema Parsing}: Then, the SL is implemented based on a simple principle: 
\textit{The schema information mentioned in the \texttt{PreSQL} is concise and relevant to the question}, and thus, the table/column entities in the statement can be directly parsed as the linking results. 

Rather than designing strategies such as sorting  \cite{c3} or extraction  \cite{din-sql} to make LLM output relevant database references, it is better to generate complete \texttt{PreSQL} statements directly. \re{It is also proven in Table \ref{tab:sl_comp}, via such a \texttt{PreSQL} pipeline, our execution accuracy can be boosted by as high as \textbf{+12.6\%}, compared with the traditional methods that rely on LLM to directly link schema}. 

There are three advantages: (1) Most LLMs are pre-trained with Text2SQL or SQL-related corpus, so presumably, their ability on SQL generation is stronger than the schema linking output (e.g., a list with \texttt{table.column} format) following instructions. {(2) Generating SQL is more generic and applicable for coding-specific LLMs such as CodeLlama \cite{codellama}. (3) The execution accuracy is theoretically monotonic. For a correctly executed \texttt{PreSQL}, the linked results are always correct. Therefore, the resulting final SQL (\texttt{FinSQL}) should also be correct. Wrong SQL may be attributed to the LLM being confused by the verbose prompt, and there is room to be correct with a concise and simplified prompt.} It should be noted that column linking has poor fault tolerance and may be inapplicable to weak LLMs, so we only perform table linking at this stage.

\subsection{SQL Generation: To Generate the Final SQL (\texttt{FinSQL})}
\label{sec:finsql}
\textbf{The Final Prompt}: After the schema linking, the mentioned tables and columns in \texttt{PreSQL} are utilized to reorganize and simplify the $RE_p$ prompt. Specifically, all the contexts that are irrelevant to the linked tables or columns in the prompt are removed adaptively, including the schema properties, database references, and foreign key declarations: \re{e.g., in Listing 2, if only the table `\texttt{singer}' is linked and the table `\texttt{singer\_in\_concert}' not, then all the lines related to the table `\texttt{singer\_in\_concert}' (Line 5, 10-14) can be pruned, as shown in Listing 4 in Appx.  \ref{ill_prompt}.} With the parsed schemas, the prompt is much simplified. \re{According to our statistics, our tokens in prompt can be saved as much as \textbf{31\%} after the pruning of \texttt{PreSQL}-schema linking.} 

\textbf{SQL Generation}: \re{To summarize, we first feed full schema information (all the tables and columns) following the Listing \ref{list:3shotrd} into a strong LLM (e.g., GPT-4) to obtain a \texttt{PreSQL}, and parse the \texttt{PreSQL} for schema linking. 
Then, the prompt is simplified by pruning irrelevant schemas and used to instruct one LLM or multiple LLMs} to generate the \texttt{FinSQL}. \re{We will introduce how to orchestra multi-LLMs.}

\subsection{Post-calibration: Cross-Consistency Is Better than Self-Consistency}
\label{sec:cc}

As a canonical trick, the self-consistency approach is widely employed for post-refining the executed results, which is conducted by increasing the randomness of an LLM output at high temperatures to generate diverse results and followed by majority voting among multiple executed results. However, it has been reported that high temperatures may hurt the performance of LLMs (increasing model hallucinations)  \cite{renze2024effect}, especially for deterministic tasks (e.g., coding). Besides, the diversity of a single model is usually limited.
Instead of conducting self-consistency, we propose instructing several LLMs under lower temperatures to generate stable and high-quality SQLs and then casting votes across the SQLs' executed results. 
Such a cross-consistency strategy can diversify the SQL queries and maintain LLMs' performance as low temperatures are set up.

In terms of voting, we propose 
two feasible implementation strategies for cross-consistency: naive voting across several LLMs and difficulty-aware voting according to PreSQL complexity.

\textbf{Naive voting across several LLMs}: 
Then, the full schema prompt is simplified and used to instruct several LLMs, each of which generates a \texttt{FinSQL} with a low temperature. 
All the \texttt{FinSQL}s and the re-used \texttt{PreSQL} are executed with the SQL database engine, and the queried results are obtained.
Finally, we check the results of each SQL execution and take the majority of the results as the final answer. \re{The \texttt{FinSQL} will be selected randomly, whichever yields the final answer.}

\textbf{Difficulty-aware voting according to \texttt{PreSQL} complexity}: In addition to the naive voting strategy, we further refine the voting rules based on the difficulty of \texttt{PreSQL}. 
As different LLMs specialize in questions with different complexity grades, putting them together in a voting pool can produce biased results. 
The complexity of questions is classified into four grades (i.e., easy, medium, hard, and extra) by the abstract syntax trees of \texttt{PreSQL} according to  \cite{zhong_test_suit}, and each grade of questions is solved by a distinct set of candidate LLMs for voting.
With such fine-grained voting, we can maximize the potential of LLMs and significantly mitigate the voting bias.

\vspace{-5pt}
\section{Experiments}
\label{sec:others}
\subsection{Experimental Setup}
\vspace{-5pt}
\textbf{Dataset and Metrics}: 
We evaluate our PET-SQL in Spider  \cite{spider} benchmark, which is a large-scale cross-domain Text2SQL dataset. It contains 8659 instances in training split and 1034 instances in development split over 200 databases, with non-overlapping databases in each set, and 2147 instances are holdout as the test set across another 34 databases. 
The evaluation of Text2SQL performance of methods is conducted by execution accuracy (EX), following the official test-suit\footnote{\url{https://github.com/taoyds/test-suite-sql-eval}} defined in  \cite{zhong_test_suit}. 
Additionally, we also conduct experiments on the Bird-SQL \cite{bird} dataset to evaluate the effectiveness and generalization of our methods, with the results reported in Appendix.
EX measures the proportion of questions in the evaluation set where the execution results of both the predicted and ground-truth inquiries are identical, relative to the total number of queries.

\textbf{Evaluated LLMs}: Five LLMs are used to validate the superiority of our PET-SQL, and we report the best results of PET-SQL on the test set, which can surpass rank 2 with about 1\% improvement on the Spider leaderboard. 
The LLMs are: CodeLlama-34B  \cite{codellama}, SQLCoder-34B  \cite{sqlcoder}, InternLM-70B  \cite{2023internlm}, SenseChat-70B  \cite{sensechat}, and GPT-4 (version on 2023.06.13)  \cite{achiam2023gpt}.
The first two are specific to coding, especially on SQL, while the others are generic LLMs. Implementation is detailed in Appx.  \ref{impdetail}.

\vspace{-5pt}
\subsection{Overall Performance}
\vspace{-5pt}
\subsubsection{Spider Leaderboard}
\vspace{-5pt}
\begin{wraptable}{r}{0.55\textwidth}
\vspace{-20pt}
\centering
\caption{The performance on Spider leaderboard}
\label{tab:perf}
\resizebox{0.55\textwidth}{!}{
\begin{tabular}{ll}
\toprule
\textbf{Methods}                    & \textbf{EX}    \\\midrule
RESDSQL-3B + NatSQL \cite{li2023resdsql} &79.9\%\\
C3 + ChatGPT + Zero-Shot   \cite{c3}            &82.3\%\\
DIN-SQL + GPT-4   \cite{din-sql}          & 85.3\%  \\
DAIL-SQL + GPT-4 + Self-consistency  \cite{dail-sql}         & 86.6\% \\ \midrule
PET-SQL (Naive voting)        & \underline{86.7\%} \\
PET-SQL (Difficulty-aware voting) & \textbf{87.6\%} \\\bottomrule
\end{tabular}}
\vspace{-10pt}
\end{wraptable} 

In Table \ref{tab:perf}, we report the performance of our method and baselines on the Spider test dataset. 
For naive voting, the executed results from the above five LLMs and the \texttt{PreSQL} are voting with equal weights, while the selection strategy of LLMs in difficulty-aware voting is illustrated in Table \ref{tab:difficulty-aware}.
Since the leaderboard is closed and rejects new submissions, we tested the performance of our approach offline, with the official test-suit.
Thus, it can be concluded that our method achieves the highest execution accuracy among all non-learning-based methods.
Specifically, our method surpasses DAIL-SQL by 1\% and achieves the best among the open-source methods in the leaderboard \footnote{Since the 1st-rank MiniSeek is not publicly available, we do not engage it in comparisons.}. Even with the naive voting strategy, the superiority of our PET-SQL still holds.


\vspace{-5pt}
\subsubsection{Comparison under other foundation LLMs}
\vspace{-5pt}
\re{We further} validate the performance difference and consistency between our PET-SQL and
other top 2 methods (DAIL-SQL, \re{DIN-DQL}) \re{when using each different foundation LLM}.

\begin{wraptable}{r}{0.75\textwidth}
\centering
\vspace{-20pt}
\caption{\re{Comparing top 3 methods} under a single LLM (9-shot), \re{thus our scheme linking (SL) and cross-consistency (CC) are removed.}}
\label{tab:comp_DAIL-SQL}
\resizebox{0.75\textwidth}{!}{
\begin{tabular}{c|c|ccc}
\toprule
        \textbf{Datasets}              &   \textbf{Methods}       & \textbf{{CodeLlama-34B}} & \textbf{SQLCoder-34B} & \textbf{InternLM-70B}  \\\midrule
\multirow{3}{*}{Spider-dev}
& DIN-SQL \cite{din-sql} &  0.673     & 0.678     & 0.686               \\\cmidrule(l){2-5} 
& DAIL-SQL \cite{dail-sql} & \textbf{0.756}   & \textbf{0.709}        & \underline{0.742}              \\\cmidrule(l){2-5} 
                      & PET-SQL \re{(w/o SL, CC)}     & \underline{0.750}     & \textbf{0.709}      & \textbf{0.750}              \\\midrule
\multirow{3}{*}{Spider-test} 
& DIN-SQL \cite{din-sql} & 0.638     & 0.626       & 0.672             \\\cmidrule(l){2-5} 
& DAIL-SQL \cite{dail-sql} & \underline{0.720}       & \underline{0.697}        & \underline{0.707}          \\\cmidrule(l){2-5} 
                      & PET-SQL \re{(w/o SL, CC)}     & \textbf{0.744}    & \textbf{0.741}      & \textbf{0.714}             \\\bottomrule 
\end{tabular}}
\vspace{-5pt}
\end{wraptable}
\textbf{\re{Our general framework and SQL-tailored prompt are universally effective for various LLMs, even without schema linking and cross-consistency}}: In this setting,  \re{only one LLM (\re{specified in each column}) attends to the Text2SQL task during the comparison}. Thus, we remove \re{our} schema linking and cross-consistency modules in the framework since their implementations involve GPT-4 and other LLMs, which may cause unfairness. The results are shown in Table \ref{tab:comp_DAIL-SQL}. Generally, even without schema linking and cross-consistency, the performance of our approach and DAIL-SQL are mutually exclusive on the LLMs on the dev set. While on the test set, our approach scores an overwhelming dominance, achieving 1\%$\sim$6\% improvements against the DAIL-SQL. All of these verify the superiority of PET-SQL's general framework and prompt design.

\vspace{-5pt}
\subsection{A Deep Dive into Prompt Design}
\vspace{-5pt}
\textbf{\re{Our few-shot $RE_p$ prompt is better than code representation and OpenAI demonstration}}: 
We compare our few-shot $RE_p$ with $CR_p$ and $OD_p$, which are recommended by  \cite{dail-sql}. It should be noted that the selected question-SQL demonstrations in them are the same. They are evaluated on the Spider development and test datasets separately, with \re{the same selected question-SQL demonstrations in them}. As shown in Table \ref{tab:pmp_perf} in Appx.  \ref{proposedprompt}, we recap the conclusions here: our proposed prompt $RE_p$ demonstrates superior performance under the zero-shot setting \re{across three different datasets and three different foundation LLMs}, with roughly 1\% to 7\% improvements compared to $CR_p$. This indicates the versatility of $RE_p$, whether for coding-specific or generic LLMs.

\begin{wraptable}{r}{0.65\textwidth}
\centering
\vspace{-20pt}
\caption{The ablation of the $RE_p$ on Spider-dev set (0-shot): \re{the full  $RE_p$ highlighted in grey, the most important component in boldface (lowest EX), the second underlined.}}
\label{tab:ab_pmp}
\resizebox{0.65\textwidth}{!}{
\begin{tabular}{c|ccc}
\toprule
            & \textbf{CodeLlama-34B} & \textbf{SQLCoder-34B} & \textbf{InternLM-70B }   \\\midrule
$RE_p$ & \cellcolor{gray!25} 73.80\%   & \cellcolor{gray!25} 65.70\%  & \cellcolor{gray!25} 70.70\% \\\midrule
w/o OR      & 71.50\%   & 64.60\%   & \underline{69.90\%}     \\\midrule
w/o CV      & \underline{71.30\%} & \textbf{61.70\%}    & \textbf{69.10\%}   \\\midrule
w/o FK      & \textbf{66.30\%}  & \underline{62.30\%}   & 71.60\%  \\
\bottomrule
\end{tabular}}
\vspace{-5pt}
\end{wraptable}
\textbf{Ablation of $RE_p$}: We further scrutinize the effects of three modifications in $RE_p$ compared to $OD_p$: optimization rule (OR), cell values reference (CV), and foreign key declarations (FK) \re{by ablation studies under the zero-shot setting.} 
\re{As shown in Table \ref{tab:ab_pmp}},
(1) \textbf{\re{Our cell values reference is the most important}}: Discarding it in InternLM and SQLCoder leads to \textbf{-2.4\%} and \textbf{-6.4\%} decreases in EX. This indicates that ensuring the presence of sampled cell values is crucial for LLM generation, and with these references, LLM can better navigate the database and formulate accurate queries. This reduces ambiguity and improves the efficiency of the generated SQL queries. (More results can be referred to in Table \ref{tab:ab_pmp2} in Appx.  \ref{ablonbird}) 
(2) \textbf{\re{Optimization rule also always helps for all three models}}: Dropping OR also shows a consistent impact on all three models' performances (\textbf{-3\%}). (3) \textbf{\re{Foreign key instead has two-sided impact}}: In CodeLlama, it boosts the performance to the most extent by \textbf{+7.5\%}, but in InternLM, not using FK is surprisingly better. 
To conclude, the CV has the greatest significance, followed by the OR and FK components. 

\vspace{-5pt}
\subsection{A Deep Dive into Schema Linking}
\vspace{-5pt}
The linked schema is parsed from the \texttt{PreSQL}, which is generated by GPT-4 with full schema information. To validate its effect, we introduce the table recall metrics $R_e$ and $R_s$ (higher is better, defined in Appx.  \ref{tabrec}), and compare the EX of \texttt{FinSQL}s generated by the LLMs based on simplified and unsimplified schemas separately.

\newcolumntype{C}[1]{>{\centering\arraybackslash}p{#1}}
\begin{table}[h]
\centering
\vspace{-5pt}
\caption{\re{Comparing schema linking (SL) performance (based on table recall metrics)} on Spider dataset between Demo-fixed SL and \texttt{PreSQL}-based SL}
\label{tab:sl_comp}
\resizebox{0.9\textwidth}{!}{
\begin{tabular}{c|c|C{1.1cm}|C{1.1cm}|C{1.1cm}|C{1.1cm}|C{1.1cm}|C{1.1cm}}
\toprule
\multirow{2}{*}{\textbf{Datasets}}   & \multirow{2}{*}{\textbf{SL Methods}}   &  \multicolumn{2}{c}{\textbf{CodeLlama-34B}} & \multicolumn{2}{c}{\textbf{SQLCoder-34B}} & \multicolumn{2}{c}{\textbf{InternLM-70B}}  \\\cmidrule(l){3-8}
                      & & $R_e$             & $R_s$             & $R_e$             & $R_s$            & $R_e$             & $R_s$            \\\midrule
\multirow{2}{*}{{Spider-dev}}  & Demo-fixed SL &  0.804         & \textbf{0.935}         &  0.815               &      0.946     &  0.676         & \textbf{0.941}                         \\\cmidrule(l){2-8}
  & \texttt{PreSQL}-based SL &   \textbf{0.864}            &    0.931     &        \textbf{0.856}         &   \textbf{0.967}           &   \textbf{0.801}              &     0.846                      \\\midrule
\multirow{2}{*}{{Spider-test}}  & Demo-fixed SL &  0.802         & 0.927      &  0.790               &     0.919     &  0.701         & \textbf{0.965}                            \\\cmidrule(l){2-8}
  & \texttt{PreSQL}-based SL &   \textbf{0.827}            &    \textbf{0.935}      &        \textbf{0.840}         &   \textbf{0.943}         &   \textbf{0.827}              &     0.932                      \\\bottomrule
\end{tabular}}\vspace{-5pt}
\end{table}

\begin{wraptable}{r}{0.7\textwidth}
\centering
\vspace{-15pt}
\caption{The performance with and without SL  (9-shot)}
\resizebox{0.7\textwidth}{!}{
\begin{tabular}{c|cccc}
\toprule
        & \textbf{Codellama-34B} & \textbf{SQLCoder-34B} & \textbf{InternLM-70B} & \textbf{GPT-4}\\\midrule
w/o SL  & 74.4\%          & 74.1\%    & 71.4\%     & 85.2\%   \\\midrule
w/ SL & \textbf{78.2\%}        & \textbf{76.9\%}     & \textbf{75.4\%}       & \textbf{85.5\%} \\\bottomrule 
\end{tabular}}
\label{tabsl}
\vspace{-10pt}
\end{wraptable}
To evaluate the superiority of the proposed \texttt{PreSQL}-based SL, in Table \ref{tab:sl_comp}, we compare it with the SL 
under demo-fixed prompt, which is used in DIN-SQL. It becomes evident that (1) \textbf{the proposed SL
 can significantly extract useful information}. \re{(2) \textbf{It also significantly shortens the prompts.} By using the linked schema, we can cut the average token length of prompt from 924.2 to 630.8, by more than \textbf{33\%}.}

With the linked tables, we further simplify the prompt and then evaluate the performance of LLMs on the test set with the 9-shot setting. The results are demonstrated in Table \ref{tabsl}. 
(1) \textbf{The simplified prompt has improved performance across all LLMs} achieving 1\%$\sim$6\% improvements, especially for CodeLlama, which validates the effectiveness of our \texttt{PreSQL}-based approach. (2) \textbf{Smaller LLMs tend to benefit significantly from SL while larger LLMs gain slightly.} This is likely due to the inherent differences in their capacities and parameters for retrieving useful information from raw text.
In conclusion, SL is an effective technique that enhances the performance and versatility of LLMs by summarizing structured information.
\vspace{-5pt}
\subsection{A Deep Dive into Cross-consistency}
\vspace{-5pt}
\begin{wraptable}{r}{0.55\textwidth}
\vspace{-10pt}
\centering
\caption{The performance of cross consistency (9-shot)}
\label{tab:cc}
\resizebox{0.55\textwidth}{!}{
\begin{tabular}{ccc}
\toprule
Methods                    & \thead{\textbf{Spider-dev}\\(w/o GPT-4)} & \thead{\textbf{Spider-test}\\(w/ GPT-4)}  \\\midrule
Baseline (best single LLM)              & 75.3\%          & 85.5\% \\
+ Self-consistency    &       76.1\%          & -       \\
+ CC (Naive voting)        & \textbf{82.2\%}          & 86.7\% \\
+ CC (Difficulty-aware voting) & -              & \textbf{87.6\%} \\ \bottomrule
\end{tabular}}
\vspace{-10pt}
\end{wraptable}

In Table \ref{tab:cc}, we report the performance of cross-consistency (CC) on dev and test sets. 
Considering the token cost, we give up the SQL generation by GPT-4 on the dev set, and the best result from the remaining LLMs is chosen as the baseline (i.e., CodeLlama). For comparison, we also perform self-consistency with CodeLlama (i.e., calling it 5 times, $\textit{temperature}=0.5$). 

It can be observed that, 
(1) \textbf{Cross-consistency is much better than self-consistency due to diversity}.~
On the dev set, the gain of self-consistency is trivial (only 0.8\%), while the cross-consistency with naive voting achieves 6.9\% improvements. On the test set, the naive voting of CC can already improve +1.2\%, while In DAIL-SQL \cite{dail-sql}, self-consistency (five SQLs generated by GPT-4) only achieves +0.4\%. \re{Thus, bringing diverse LLMs can better boost the accuracy since different LLMs may make different errors, and letting them vote mitigates each other's mistakes.} 
(2) \textbf{Difficulty-aware voting is a better implementation of CC}. Compared to naive voting, the difficulty-aware voting strategy can further exploit the potential of LLMs and significantly mitigate the voting bias, improving the performance from 86.7\% to 87.6\% on the test set.

\begin{wraptable}{r}{0.7\textwidth}
\centering
\vspace{-15pt}
\caption{\re{Suggested set of LLMs for difficulty-aware voting w.r.t various difficulty questions on Spider-test}}
\label{tab:difficulty-aware}
\resizebox{0.7\textwidth}{!}{
\begin{tabular}{llc}
\toprule
\textbf{Difficulty}      & \textbf{Models for CC}                               & \textbf{EX}    \\\midrule
Easy   & SQLCoder-34B, InternLM-70B, SenseChat-70B           & 0.932 \\
Medium & GPT-4, InternLM-34B, SenseChat, CodeLlama-34B & 0.907 \\
Hard   & GPT-4, InternLM-70B, SenseChat-70B            & 0.849 \\
Extra  & GPT-4, SenseChat-70B                              & 0.759 \\\midrule
\textbf{Total}  &                                            & 0.876 \\\bottomrule
\end{tabular}}\vspace{-5pt}
\end{wraptable}
\textbf{For questions in different difficulty levels, a tailored set of multi-LLMs can further optimize the accuracy}. Table \ref{tab:difficulty-aware} gives the details of our recommended set of LLMs for each difficulty level and their EX on the test set
using the \texttt{PreSQL} only. (1) \textbf{Smaller LLMs are sufficient to handle the easy questions}: For easy questions, smaller LLMs can achieve 93.2\% EX, demonstrating their efficacy in handling less complex questions. (2) \textbf{For difficult questions, there is a clear need for more powerful LLM} (i.e., GPT-4) to maintain high performance. This performance disparity underscores the importance of selecting the appropriate LLMs based on the difficulty of the questions.


\begin{wraptable}{r}{0.3\textwidth}
\centering
\vspace{-15pt}
\caption{\re{CC as plug-and-play}}
\label{tab:cc-transfer}
\resizebox{0.28\textwidth}{!}
{
\begin{tabular}{c |c} 
\toprule
        &  EX \\\midrule
DAIL-SQL  & 0.720  \\\midrule
DAIL-SQL + CC    & 0.735  \\\bottomrule
\end{tabular}}\vspace{-10pt}
\end{wraptable}
\textbf{CC as a plug-and-play can always boost performance}: CC can serve as a plug-and-play module for other SOTA methods. We evaluate its performance with DAIL-SQL on the Spider-test set as shown in Table \ref{tab:cc-transfer}. 
With the introduction of CC, the DAIL-SQL can be further enhanced by \textbf{+2\%} improvements. This suggests that the CC strategy is still effective on other methods. CodeLlama is used as the single LLM setting, and then SQLCoder and InternLM are added as the CC setting. 

\re{
\textbf{More Experiments in Appendix}: We also provide more experimental results to evaluate the superiority or effectiveness of our proposed methods. 
Conducted on Bird-SQL dataset \cite{bird}: (1) the comparison between our PET-SQL and other SOTA methods (that is, DIN-SQL \cite{din-sql} and DAIL-SQL \cite{dail-sql}) under different foundation LLMs are shown in Appx. \ref{valbird}; (2) The ablation results of the proposed prompt $RE_p$ are further reported in Appx. \ref{ablonbird}. Furthermore, we compare $RE_p$ with two well-defined prompts from DAIL-SQL \cite{dail-sql} (i.e., $CR_P$ and $OD_p$) under the zero-shot setting, as detailed in Appx. \ref{proposedprompt}. In Appx. \ref{deepsl}, we also analyze the SL results by the \texttt{PreSQL} from GPT-4. Finally, the significance of GPT-4 in CC is underscored in Appx. \ref{ccgpt4}.
}
\vspace{-5pt}
\section{Conclusion}
\vspace{-10pt}
This paper presents a two-round framework based on pre-trained LLMs, PET-SQL, which aims to address challenges in Text2SQL tasks by enhancing the prompt and leveraging cross-consistency across LLMs. Our approach achieves 87.6\% execution accuracy on the Spider leaderboard.
We also propose a \texttt{PreSQL}-based schema linking method to simplify prompt information and improve the efficiency and accuracy of LLMs in generating SQL queries. 
Overall, the PET-SQL framework demonstrates promising results and opens avenues for further advancements in Text2SQL tasks. 
\bibliographystyle{unsrt}  
\bibliography{templateArxiv}  

\appendix

\section{Appendix}
\subsection{Implementation details}\label{impdetail} To decrease the randomness of LLMs output, the temperatures are set extremely low ($10^{-7}$ for all the non-OpenAI LLMs and 0 for GPT-4). The max lengths of input and output tokens are 4096 and 200, respectively. 
In the question skeleton-based demonstration retrieval, we traverse all the database schema and check if the table or column names appear in questions. The matched entities will be masked by \texttt{<mask>} token. Based on the similarities between the sentence embedding of the target and candidate question skeletons, top-9 similar question-SQL pairs in the demonstration pool are selected and then constructed as the in-context. For cross-consistency, the executed results from the above five LLMs and the \texttt{PreSQL} are voting with equal weights.
\subsection{The definition of table recall metrics} \label{tabrec}
To evaluate the effect of schema linking results, we introduce the table recall metrics, which are two-fold and defined as
\begin{itemize}
    \item {Totally Equal:} $R_{e} = {\sum_{i=1}^N \mathbf{1}_e}/{N}$, where $N$ is the number of instances in the test set. $\mathbf{1}_e$ is an indicator function, which returns 1 if the linked tables are exactly the tables that appeared in ground-truth SQL (GT tables) and 0 otherwise. $R_{e}=1$ is the ideal schema linking result, which means all the question-related tables are recalled, and there is no verbose information. (higher is better)
    \item {Subset:} $R_{s} = {\sum_{i=1}^N \mathbf{1}_s}/{N}$. $\mathbf{1}_s$ is an indicator function, which returns 1 if GT tables are the subset of the linked tables and 0 otherwise. $R_{s}$ represents the upper limit of the accuracy that LLM can achieve with schema linking. Even if the linked table is not sufficiently simplified, useful tables are still retained in the prompt.
\end{itemize}

\subsection{The illustration of the simplified prompt with the schema linking results}\label{ill_prompt}
As shown in Listing 4, the initial prompt includes several entities (e.g., ``\texttt{singer\_in\_concert}'') that are not directly relevant to the primary question at hand, where the strikethrough (in lines 5, 10-14) indicates removing contexts in $RE_p$ via schema linking. By applying schema linking, we can streamline the prompt by eliminating unnecessary information and focusing on the critical elements that directly support the question ``\texttt{How many singers do we have?}''.
\setcounter{lstlisting}{3}
\begin{lstlisting}[language=Prompt, label={lst:simplification}, caption={A simplified prompt when the linked table from PreSQL is ``singer''. The strikethrough (in lines 5, 10-14) indicates removing contexts.}, float=htbp]
### Some example pairs of questions and corresponding SQL queries are provided based on similar questions:

### How many farms are there?
SELECT count(*) FROM farm

### How many books are there?
SELECT count(*) FROM book

### How many actors are there?
SELECT count(*) FROM actor

### Answer the question by SQLite SQL query only and with no explanation. You must minimize SQL execution time while ensuring correctness.
### Sqlite SQL tables, with their properties:
#
# singer(Singer_ID,Name,Country,Song_Name,Song_release_year,Age,Is_male);
|# singer_in_concert(concert_ID,Singer_ID);|
#
### Here is some data information about database references.
#
# singer(Singer_ID[1,2,3],Name[Joe,Timbaland,Justin Brown],Country[Netherlands,United States,France],Song_Name[You,Dangerous,Hey Oh],Song_release_year[1992,2008,2013],Age[52,32,29],Is_male[F,T,T]);
|# singer_in_concert(concert_ID[1,1,1],Singer_ID[2,3,5]);|
|#|
|### Foreign key information of SQLite tables, used for table joins:|
|#|
|# singer_in_concert(Singer_ID) REFERENCES singer(Singer_ID);|
# 
### Question: How many singers do we have?
### SQL: 
\end{lstlisting}
\subsection{The validation on the Bird-SQL dataset}\label{valbird}
Additionally, we have also conducted experiments on the bird develop (bird-dev) dataset \cite{li2024can}, recognized for its heightened complexity. Table \ref{tab:comp_DAIL-SQL_bird} presents the outcomes of SOTA methods with various foundational LLMs. The DAIL-SQL method stands out, exhibiting superior performance across all evaluated metrics on the Bird-dev dataset. Our proposed PET-SQL method performs admirably, particularly excelling in the SQLCoder with a score of 0.391.
\begin{table}[h]
\centering
\caption{The comparison under foundation LLMs on Bird-SQL dataset (9-shot)}
\label{tab:comp_DAIL-SQL_bird}
{
\begin{tabular}{c|c|ccc}
\toprule
        \textbf{Datasets}              &   \textbf{Methods}       & \textbf{{CodeLlama}} & \textbf{InternLM} & \textbf{SQLCoder} \\\midrule
\multirow{3}{*}{Bird-dev}
& DIN-SQL & 0.286        & 0.317        & 0.301        \\\cmidrule(l){2-5} 
& DAIL-SQL & \textbf{0.399}         & \textbf{0.331}        & 0.382        \\\cmidrule(l){2-5} 
                      & PET-SQL(Ours)     & 0.348         & 0.274        & \textbf{0.391}        \\\bottomrule 
\end{tabular}}
\vspace{-10pt}
\end{table}
\vspace{-5pt}
\subsection{The ablation of $RE_p$ on Bird-dev}\label{ablonbird}
\vspace{-5pt}
To verify the effectiveness and generalization of the proposed $RE_p$, we also conduct the ablation study on the Bird-dev dataset. The results are shown in Table \ref{tab:ab_pmp2}. 
The conclusion is the same as the one summarized in Table \ref{tab:ab_pmp}:
The performance of all three LLMs (CodeLlama-34B, InternLM-70B, and SQLCoder-34B) decreases when any of the components (OR, CV, FK) are removed, which demonstrates that each component of $RE_p$ contributes to the LLMs' ability to generalize and perform well on the Bird-dev dataset. 

\begin{table}[h]
\centering
\caption{The ablation of the $RE_P$ on Bird-dev set (0-shot)}
\label{tab:ab_pmp2}
{
\begin{tabular}{c|ccc}
\toprule
            & \textbf{CodeLlama-34B} & \textbf{InternLM-70B }   & \textbf{SQLCoder-34B} \\\midrule
$RE_p$ & \textbf{35.50\%}   & \textbf{30.80\%} &  \textbf{34.50\%} \\\midrule
w/o OR      &  \underline{34.16\%}   &  \underline{30.31\%} &  \underline{33.70\%}  \\\midrule
w/o CV      &  33.18\%   &  29.73\% &  31.29\%    \\\midrule
w/o FK      &  32.14\%   &  30.05\% &  {33.12\%}  \\
\bottomrule
\end{tabular}}
\end{table}

\subsection{Further analysis of the proposed prompt $RE_p$}\label{proposedprompt}
To further illustrate the effectiveness of the proposed prompt, we compared it with two well-defined prompts from DAIL-SQL \cite{dail-sql}, namely $CR_p$ and $OD_p$, under the zero-shot setting. The results are presented in Table \ref{tab:pmp_perf}. It is evident that even with different foundation LLMs, our proposed $RE_p$ remains effective, achieving about 
a 7\% to 10\% relative performance boost across various datasets compared to the baseline prompts. This improvement signifies the robustness and generalizability of our proposed prompt $RE_p$ in enhancing model performance under the 0-shot setting.
\begin{table}[h]
\centering
\caption{The performance comparison between different prompt styles  (0-shot)}
\label{tab:pmp_perf}
\begin{tabular}{c|c|ccc}
\toprule
          \textbf{Datasets}                 &   \textbf{Prompting}     & \textbf{Codellama-34B} & \textbf{SQLCoder-34B}& \textbf{InternLM-70B}  \\ \midrule
\multirow{2}{*}{Bird-dev}  & $CR_p$ & 0.333    & 0.266     & 0.265         \\\cmidrule(l){2-5} 
                          & $RE_p$ & \textbf{0.355}    & \textbf{0.345}      & \textbf{0.308}         \\\midrule
\multirow{2}{*}{Spider-dev}  & $CR_p$ & 0.685       & 0.562    & 0.701         \\\cmidrule(l){2-5} 
                          & $RE_p$ & \textbf{0.738}      & \textbf{0.657 }      & \textbf{0.707}       \\\midrule
\multirow{3}{*}{Spider-test} & $CR_p$ & \underline{0.694}      & \underline{0.631}      & 0.649         \\\cmidrule(l){2-5} 
                          & $OD_p$ & 0.609       & 0.579    & \underline{0.673}          \\\cmidrule(l){2-5} 
                          & $RE_p$ & \textbf{0.717}    & \textbf{0.678}       & \textbf{0.699}          \\ \bottomrule
\end{tabular}
\end{table}

\subsection{Further analysis of the PreSQL-based schema linking}\label{deepsl}
\begin{wraptable}{r}{0.4\textwidth}
\centering
\vspace{-10pt}
\caption{The metrics of SL with GPT-4}
\label{tab:recall}
\begin{tabular}{ccc}
\toprule
LLM for SL & $R_e$     & $R_s$     \\\midrule
GPT-4    & 0.94 & 0.98 \\\bottomrule
\end{tabular}
\vspace{-10pt}
\end{wraptable}

Table \ref{tab:recall} shows the schema linking (SL for short) recall of GPT-4 (i.e., \texttt{PreSQL}). 
It can be seen that the recall of table linking reaches 94\% while the EX of \texttt{PreSQL} is only 85.2\%, and the limitations of LLMs may cause the gap. Prompt simplification with linked tables enables LLM to correct the wrong \texttt{PreSQL} cases.

\begin{wraptable}{r}{0.45\textwidth}
\centering
\vspace{-10pt}
\caption{The average of tables in prompts}
\label{tab:tabnum}
\begin{tabular}{ccc}
\toprule
w/o SL & w/ SL     & Ground truth     \\\midrule
4.89    & 1.60 & 1.57 \\\bottomrule
\end{tabular}
\vspace{-8pt}
\end{wraptable}

In Table \ref{tab:tabnum}, we further summarize the average tables mentioned in the prompt and the ground truth (i.e., the average of GT tables). In the full schema (without SL), the average number of tables is 4.89, while in the simplified schema (with SL), it is 1.60 (close to GT). This means that SL significantly simplifies useless table information in prompts.
\vspace{-5pt}
\subsection{Further analysis of the cross consistency} \label{ccgpt4}
\vspace{-5pt}
GPT-4's pivotal role is uncovered in enhancing the final outcomes. We conduct experiments to gauge its significance. Table \ref{tab:imp_gpt} demonstrates the importance of GPT-4 within the Spider-test set:
Specifically, utilizing GPT-4 alone yielded an accuracy of 85.50\%, showcasing its substantial impact. Conversely, excluding GPT-4 from the other four models used for CC led to a noticeable decline in performance, resulting in an accuracy of 79.80\%. Replacing GPT-4 with a weaker model (InternLM2-20B) within the CC framework further exacerbated this decrement, yielding an accuracy of 79.08\%. However, employing all five models within the CC framework resulted in a notable enhancement, achieving an accuracy of 87.60\%. These findings underscore GPT-4's indispensable contribution to the overall performance, illustrating its pivotal role in enhancing CC. 
\begin{table}[h]
\centering
\caption{The importance of GPT-4 on Spider-test set}
\label{tab:imp_gpt}
\begin{tabular}{ll}
\toprule
\textbf{Models}                                  & \textbf{EX} \\\midrule
GPT-4 Only                          &  85.50\%  \\
The other four models w/o GPT-4  for CC  & 79.80\%   \\
Rep. GPT-4 with a weak model for CC  &   79.08\% \\
All the five models for CC        &  87.60\% \\\bottomrule
\end{tabular}
\end{table}
\end{document}